\documentclass[10pt,journal]{IEEEtran}
\IEEEoverridecommandlockouts
\ifCLASSOPTIONcompsoc
  \usepackage[nocompress]{cite}
\else
  \usepackage{cite}
\fi

\ifCLASSINFOpdf
\else
\fi

\usepackage{cite}
\usepackage{amsmath,amssymb,amsfonts}
\usepackage{algorithmic}
\usepackage{array}
\usepackage{graphicx}
\usepackage{textcomp}
\usepackage[table]{xcolor}
\usepackage{orcidlink}
\usepackage{multirow}
\usepackage{subcaption}
\usepackage{booktabs}

\def\BibTeX{{\rm B\kern-.05em{\sc i\kern-.025em b}\kern-.08em
    T\kern-.1667em\lower.7ex\hbox{E}\kern-.125emX}}
\usepackage{listings}

\definecolor{codegreen}{rgb}{0,0.6,0}
\definecolor{codegray}{rgb}{0.5,0.5,0.5}
\definecolor{codepurple}{rgb}{0.58,0,0.82}
\definecolor{backcolour}{rgb}{0.95,0.95,0.92}

\lstdefinestyle{mystyle}{
    backgroundcolor=\color{backcolour},   
    commentstyle=\color{codegreen},
    keywordstyle = {\color{magenta}},
    keywordstyle = [2]{\color{lime}},
    keywordstyle = [3]{\color{yellow}},
    keywordstyle = [4]{\color{teal}},
    numberstyle=\tiny\color{codegray},
    stringstyle=\color{codepurple},
    basicstyle=\ttfamily\footnotesize,
    breakatwhitespace=false,         
    breaklines=true,                 
    captionpos=b,                    
    keepspaces=true,                 
    numbers=left,                    
    numbersep=5pt,                  
    showspaces=false,                
    showstringspaces=false,
    showtabs=false,                  
    tabsize=2
}

\begin{document}

\title{Semantic-Guided Natural Language and Visual Fusion for Cross-Modal Interaction Based on Tiny Object Detection  \\
}
\author{Xian-Hong Huang, Hui-Kai Su, Chi-Chia Sun, Jun-Wei Hsieh  \\
\IEEEauthorblockA{} 
\thanks{Xian-Hong Huang and Hui-Kai Su are with Department of Electrical Engineering, National Formosa University, Taiwan; 

Chi-Chia Sun is with Department of Electrical Engineering, National Taipei University, Taiwan; 

Jun-Wei Hsieh is with College of Artificial Intelligence and Green Energy, National Yang Ming Chiao Tung University, Taiwan;

Corresponding Author is Xian-Hong Huang (\textit{E-mail: 40725001@gm.nfu.edu.tw})
}
}

\maketitle
\markboth{Official Report}%
{Huang \etal{}: ParFormer}

\begin{abstract}
This paper introduces a cutting-edge approach to cross-modal interaction for tiny object detection by combining semantic-guided natural language processing with advanced visual recognition backbones. The proposed method integrates the BERT language model with the CNN-based Parallel Residual Bi-Fusion Feature Pyramid Network (PRB-FPN-Net), incorporating innovative backbone architectures such as ELAN, MSP, and CSP to optimize feature extraction and fusion. By employing lemmatization and fine-tuning techniques, the system aligns semantic cues from textual inputs with visual features, enhancing detection precision for small and complex objects. Experimental validation using the COCO and Objects365 datasets demonstrates that the model achieves superior performance. On the COCO2017 validation set, it attains a 52.6\% average precision (AP), outperforming YOLO-World significantly while maintaining half the parameter consumption of Transformer-based models like GLIP. Several test on different of backbones such ELAN, MSP, and CSP further enable efficient handling of multi-scale objects, ensuring scalability and robustness in resource-constrained environments. This study underscores the potential of integrating natural language understanding with advanced backbone architectures, setting new benchmarks in object detection accuracy, efficiency, and adaptability to real-world challenges.

\end{abstract}
\section{Introduction}
\label{sec:intro}

Object detection has been the main downstream of the image processing task. Significant advancements in object detection have been accomplished through the development of deep neural networks, including RCNN\cite{girshick2014rich, ren2016faster}, FPN\cite{lin2017feature}, SSD\cite{liu2016ssd}, and YOLO \cite{bochkovskiy2020yolov4, wang2023yolov7, wang2025yolov9, wang2024yolov10}. Despite their success, most of this method focuses on the single modality image that has limited information such as a fixed vocabulary, e.g., 80 categories in the COCO \cite{lin2014microsoft} dataset. Single modality image information with fixed categories lead to the limited the ability and applicability in the open real-world problems. 

Recent works \cite{du2022learning, gu2021open, shi2023edadet}  have explored the prevalent vision-language models \cite{jia2021scaling, radford2021learning}to address open vocabulary detection through distilling vocabulary
knowledge from language encoders, e.g., BERT \cite{kenton2019bert}. However, beside of the open vocabulary,  one of the critical challenges in computer vision is accurate recognition and localization of small objects within images. Small object detection is inherently difficult due to the limited pixel information and the influence of surrounding noise \cite{chen2019hybrid, liu2018path}. Existing object detection models often struggle with small objects, leading to suboptimal performance in applications such as surveillance, autonomous driving, and medical imaging \cite{najibi2019fa}.

Recent studies have explored cross-modal interactions to enhance small object detection by leveraging semantic information from natural language descriptions \cite{zhou2020unified, tan2019lxmert}. The integration of NLP and CV allows models to utilize contextual cues from textual data to improve visual perception tasks \cite{li2019relation}. Transformer-based architectures, such as BERT for NLP \cite{kenton2019bert} and Vision Transformers for CV \cite{dosovitskiy2020image}, have shown remarkable success in capturing long-range dependencies and contextual relationships within data, making them suitable for cross-modal applications \cite{sun2019videobert, lu2019vilbert}.

In this work, we propose a novel approach that combines the BERT language model with the Parallel Residual Bi-Fusion Feature Pyramid Network (PRB-FPN-Net) for enhanced small object detection through semantic-guided cross-modal interaction. Our method leverages the strengths of BERT in understanding natural language to extract semantic cues, which are then integrated with visual features extracted by PRB-FPN-Net, a CNN-based model optimized for small object detection \cite{chen2021parallel}. By fine-tuning both models and applying lemmatization techniques using WordNetLemmatizer \cite{gupta2020lemmachase}, we transform extracted words into their root forms to enhance the model's comprehension of natural language information. Our contributions can be summarized as follows:
\begin{itemize}
    \item Semantic-Guided Fusion: We introduce a cross-modal fusion strategy that effectively combines semantic information from text descriptions with visual features, improving the detection of small objects that are often overlooked by traditional models.
    \item Enhanced Text Encoding: By implementing advanced tokenization and lemmatization techniques, we optimize the text encoding process, enabling more accurate alignment between textual and visual modalities.
    \item Improved Performance: Experimental results on benchmark datasets, including COCO \cite{lin2014microsoft} and Objects365 \cite{shao2019objects365}, demonstrate that our model outperforms state-of-the-art methods in object detection accuracy and efficiency, particularly in resource-constrained scenarios.
    
\end{itemize}

The remainder of this paper is organized as follows. Section 2 reviews related work in cross-modal interaction and small object detection. Section 3 details our proposed method, including the integration of BERT and PRB-FPN-Net models. Section 4 presents experimental results and comparisons with existing methods. Finally, Section 5 concludes the paper and discusses future research directions.


\section{Related Work}
\label{sec:formatting}
\subsection{Conventional Object Detection}
Traditional object detection approaches generally focus on predefined categories, as illustrated by datasets like COCO \cite{lin2014microsoft} and Object365 \cite{shao2019objects365}, which contain fixed sets of 80 and 365 categories, respectively. This rigid categorization limits flexibility, as models trained on such datasets may struggle to generalize to unseen objects. Detection approaches are typically divided into region-based and pixel-based methods. Among region-based approaches, Faster R-CNN \cite{ren2016faster} introduced a two-stage framework in which a Region Proposal Network (RPN) first generates candidate regions, followed by RoI classification and regression, significantly improving detection accuracy over earlier models like R-CNN \cite{girshick2014rich}.

Single-stage models like Single Shot Detector (SSD) \cite{liu2016ssd}, You Only Look Once (YOLO) \cite{redmon2016you}, and Fully Convolutional One-Stage Object Detector (FCOS) \cite{tian2020fcos} have since been developed to improve inference speed, enabling real-time object detection. SSD and YOLO rely on predefined anchor boxes and grid-based predictions, while FCOS adopts an anchor-free approach, allowing for more flexible object localization without reliance on anchors.

The YOLO series has seen significant architectural improvements. For instance, cross-stage partial networks (CSPNet) \cite{wang2020cspnet} were introduced to reduce computational redundancy, partitioning feature maps to balance feature reuse and efficiency. This concept was extended by mixed-stage partial (MSP) networks \cite{chen2022mixed}, which combine inter-stage and intra-stage connections to enhance feature propagation. CSPNet and MSP have been incorporated into YOLO versions like YOLOv4 \cite{bochkovskiy2020yolov4} and YOLOv5 \cite{glenn_jocher_2020_4154370}, resulting in models that are both accurate and computationally efficient.

More recently, YOLOv7 \cite{wang2023yolov7} and YOLOv10 \cite{wang2024yolov10} introduced innovations in feature aggregation and a non-maximum suppression (NMS)-free approach, allowing for more accurate detection without traditional NMS, enhancing both precision and inference speed. These modifications support YOLO's use in demanding applications, such as autonomous driving and surveillance, where both accuracy and efficiency are crucial.

\subsection{Visual-Language Model}
In recent years, there has been a growing trend towards developing vision-and-language models that tackle visual recognition problems by training vision models with free-form language supervision. This approach enables models to better understand and interpret images within a broader, open-vocabulary context. Notable examples include CLIP \cite{radford2021learning} and ALIGN \cite{jia2021scaling}, which use cross-modal contrastive learning on massive datasets comprising hundreds of millions of image-text pairs. These models are capable of open-vocabulary image classification, allowing them to classify a wide range of visual concepts without being restricted to predefined categories.

Building on this foundation, ViLD \cite{gu2021open} introduced a method to adapt these vision-and-language models for zero-shot object detection. By distilling knowledge from models like CLIP and ALIGN into a two-stage object detector, ViLD achieves a breakthrough in detecting objects it has never explicitly seen during training, thus advancing zero-shot capabilities in object detection tasks. Alternatively, MDETR \cite{kamath2021mdetr} takes an end-to-end approach by training on existing multi-modal datasets that contain explicit alignments between textual phrases and corresponding objects in images. This alignment allows MDETR to perform phrase-grounding tasks effectively, linking specific phrases in text to the relevant objects in an image.

\begin{figure*}
    \centering
    \includegraphics[width=16cm]{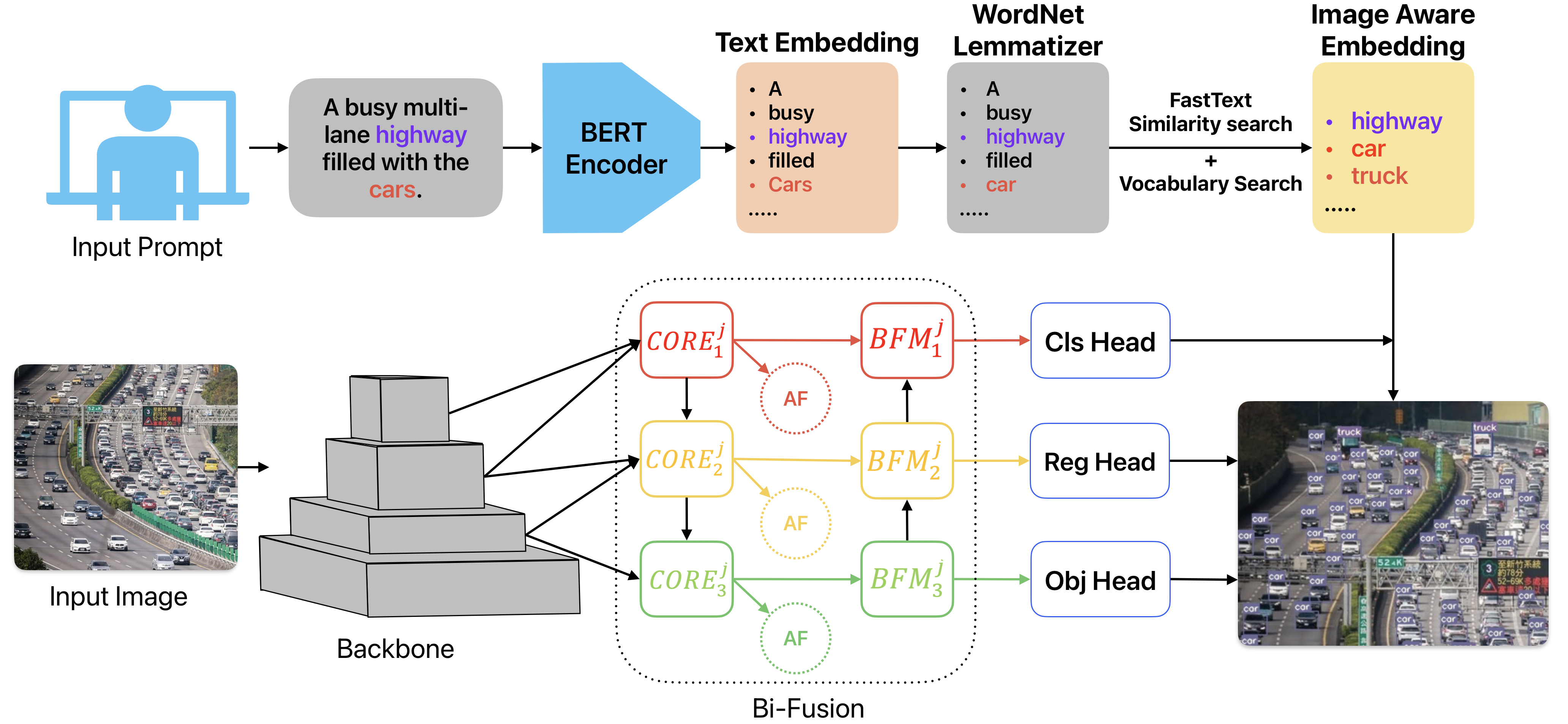}
    \caption{The proposed methodology framework outlined in this research}
    \label{fig:framework}
\end{figure*}

\section{Method}

This section details the research methodology proposed in this study, which is divided into two main parts: text encoding and image encoding. For text encoding, we utilize the BERT model \cite{devlin2018bert} \cite{devlin2019bertpretrainingdeepbidirectional} for text processing. In image object detection, the Parallel Bi-fusion Network for feature pyramid  model (PRB-FPN-Net) is employed. The PRB-FPN-Net model is modified to receive annotation information from the text encoding part, enabling it to locate corresponding entities within the input images.

\subsection{Architecture Framework}
To understand the correspondence between images and their descriptions, this study integrates two main areas: image encoding and text processing. The research employs techniques from both fields, using sentences and images as inputs to the model to generate accurate object locations. The methodology is as follows:

\begin{itemize}
    \item Text Processing: In text processing involve BERT model as the text encoder and WordLemmatizer to filter the phrase and similarity search related the class of object in the image.
\end{itemize}

\begin{itemize}
    \item Image Encoding: In image encoding, the extracted feature from the backbone pass through PRB-Net including the Bi-Fusion to keep tracks of features that are suitable to detect multi scale of object. Then in the final layer lead head including classification head, regression head, and object head related to the anchor to handle multiple scales. 
\end{itemize}

\begin{figure}[!b]
    \centering
    \includegraphics[width=0.7\linewidth]{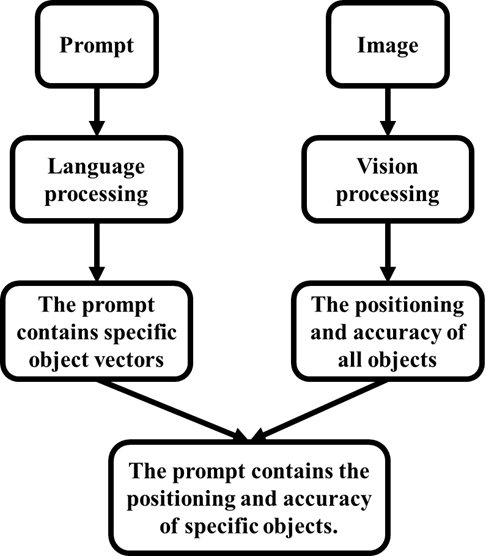}
    \caption{Overall Framework and Workflow}
    \label{fig:enter-label}
\end{figure}
To validate the proposed method, this study utilizes the BERT model for natural language processing and the PRB-FPN-Net model for image feature encoding. The research process involves collecting and preprocessing data using COCO\cite{lin2014microsoft} and Objects365\cite{shao2019objects365} image datasets. The BERT model is employed for text feature extraction, including Lemmatization with WordNetLemmatizer and vocabulary search. The  PRB-FPN-Net model is then trained on annotated image datasets for object detection. Keywords and images are integrated as inputs for keyword-based object detection, and the performance of this approach is tested for accuracy and efficiency. Finally, based on the test results, the study evaluates the method's strengths and weaknesses and optimizes the model and algorithms to enhance accuracy and applicability.

\subsection{Image Encoding}
\subsubsection{Feature Pyramid Network}
In the object detector, a robust parallel Feature Pyramid fusion design addresses the challenge of accommodating objects of various scales. This is achieved through multiple bi-fusion paths that maintain features optimal for detecting objects of all sizes, from tiny to large. Each bi-fusion path tracks scale-specific features, capturing objects at distinct sizes. Assuming there are N prediction maps, we propose running N separate, concurrent fusion paths to create N fused feature maps corresponding to each prediction map. Let $X_i$ denote the features obtained from the backbone, $BF_j$ denote the j-th BiFusion module, and $CORE^j_k$ and $BFM^j_k$ represent the Pyramidal Layer within the j-th Bi-Fusion module. The input configuration for bottom-up feature fusion can be expressed as:
\begin{equation}
    CORE^j_k = \{X_{4-k} , X_{3-k} , CORE^j_{k-1}\}
\end{equation}
where $j = 1, 2, 3$ and $k = 1, 2, 3, 4$  respectively. The input configuration for top-down feature fusion can be written as: 
\begin{equation}
BFM^j_k = {CORE^j_k , BFM^j_{k+1}}
\end{equation}
where $j = 1, 2, 3$ and $k = 1, 2, 3, 4$ respectively. By employing this hierarchical approach, our method effectively combines multi-scale features and leverages the advantages of each Bi-Fusion module, leading to improved performance on object detection and recognition tasks.
\subsubsection{Lead Head}
We utilize the lead head prediction as a guide to create coarse-to-fine hierarchical labels, which aid in training both the auxiliary and lead heads. Additionally, we introduce a novel parallelization approach for these heads, allowing more effective focus on regions of interest and capturing crucial features for object detection. This parallelization enhances feature representation by efficiently gathering information to identify and localize objects of diverse sizes without reducing efficiency. 

Within this parallelized structure, the output from each BFM module \cite{chen2021parallel} is concatenated to form a "Lead Fusion" before reaching the lead head at each level k, as shown:
\begin{equation}
 LeadFusion_k = cat \{ BFM1_k , BFM2_k , BFM3_k \} 
\end{equation}
where 
k=1,2,3,4, and the Lead Fusion Head at each level k includes the classification, regression, and object/region heads. Similarly, the output from each CORE module \cite{chen2021parallel} is concatenated to form an "Auxiliary Fusion" before the auxiliary head for each level k:
\begin{equation}
AuxFusion_k = cat\{CORE^1_k , CORE^2_k , CORE^3_k\}
\end{equation}
where k = 1, 2, 3, 4. These new designs greatly improve our model’s performance by enabling it to learn from richer training signals. A central innovation here is the parallelized structure of both lead and auxiliary heads, enhancing feature representation to capture critical information for identifying and localizing objects across scales without compromising efficiency. These advancements empower the model with robust detection capabilities, facilitating precise object identification and localization across various contexts.

\begin{table*}[!htbp]
\caption{Performance Comparison with Other Research on COCO Dataset}
\centering
\begin{tabular}{lcccccccc}
\toprule
\textbf{Model} & \textbf{AP\textsuperscript{test}} & \textbf{AP\textsubscript{50}\textsuperscript{test}} & \textbf{AP\textsubscript{75}\textsuperscript{test}} & \textbf{AP\textsubscript{s}\textsuperscript{test}} & \textbf{AP\textsubscript{m}\textsuperscript{test}} & \textbf{AP\textsubscript{l}\textsuperscript{test}} & \textbf{Parameter} & \textbf{GFLOPs} \\
\midrule
GLIP\_T(A) & 53.3\% & 71.5\% & 58.4\% & 39.1\% & 57.9\% & 66.5\% & 232 M & - \\
GLIP-T (B) & 54.0\% & 72.3\% & 59.1\% & 39.5\% & 58.5\% & 67.6\% & 232 M & - \\
GLIP-T (C) & 55.2\% & 73.5\% & 60.6\% & 40.9\% & 59.9\% & 69.1\% & 232 M & - \\
GLIP-T & 55.4\% & 73.5\% & 60.8\% & 40.4\% & 60.0\% & 69.8\% & 232 M & - \\
GLIP-L & 59.3\% & 77.4\% & 64.6\% & 44.6\% & 63.7\% & 74.2\% & 232 M & - \\
YOLO-World-v1-S & 37.4\% & 52.0\% & 40.6\% & 21.1\% & 41.4\% & 49.6\% & 12.75 M & 71.2 \\
YOLO-World-v1-M & 42.1\% & 57.0\% & 45.6\% & 26.8\% & 46.5\% & 55.3\% & 27.70 M & 131.0 \\
YOLO-World-v1-L & 45.8\% & 61.3\% & 49.8\% & 30.3\% & 46.9\% & 59.4\% & 45.32 M & 225.0 \\
YOLO-World-v2-S & 37.7\% & 52.4\% & 40.8\% & 21.4\% & 42.2\% & 50.1\% & 12.16 M & 50.8 \\
YOLO-World-v2-M & 43.1\% & 58.6\% & 46.6\% & 27.1\% & 47.8\% & 56.2\% & 27.06 M & 110.2 \\
YOLO-World-v2-L & 45.8\% & 61.5\% & 49.6\% & 30.6\% & 47.6\% & 58.5\% & 44.64 M & 203.9 \\
\midrule
Proposed-CSP & 47.2\% & 65.5\% & 51.7\% & 31.1\% & 52.0\% & 60.5\% & 58.80 M & 153.6 \\
Proposed-ELAN & 48.4\% & 66.1\% & 52.8\% & 31.1\% & 53.7\% & 62.7\% & 96.43 M & 252.8 \\
Proposed-MSP & 52.6\% & 70.2\% & 57.5\% & 35.4\% & 57.3\% & 67.1\% & 101.14 M & 368.1 \\
\bottomrule
\end{tabular}
\label{tab:coco-dataset}
\end{table*}

\begin{table*}[!ht]
\centering
\caption{Performance Comparison with Other Research on Objects365 Dataset}
\begin{tabular}{lcccccccc}
\toprule
\textbf{Model} & \textbf{AP\textsuperscript{test}} & \textbf{AP\textsubscript{50}\textsuperscript{test}} & \textbf{AP\textsubscript{75}\textsuperscript{test}} & \textbf{AP\textsubscript{s}\textsuperscript{test}} & \textbf{AP\textsubscript{m}\textsuperscript{test}} & \textbf{AP\textsubscript{l}\textsuperscript{test}} & \textbf{Parameter} & \textbf{GFLOPs} \\
\midrule
YOLO-World-v1-S & 16.3\% & 22.4\% & 17.6\% & 5.9\% & 15.5\% & 22.4\% & 157.02 M & 198.7 \\
YOLO-World-v1-M & 19.7\% & 26.1\% & 21.3\% & 7.9\% & 19.0\% & 26.9\% & 171.91 M & 280.9 \\
YOLO-World-v1-L & 24.3\% & 31.7\% & 26.4\% & 10.8\% & 24.2\% & 31.7\% & 189.60 M & 389.9 \\
YOLO-World-v2-S & 16.4\% & 22.5\% & 17.7\% & 6.0\% & 15.4\% & 22.6\% & 156.43 M & 118.0 \\
YOLO-World-v2-M & 21.0\% & 27.9\% & 22.8\% & 8.7\% & 20.4\% & 28.2\% & 171.31 M & 199.8 \\
YOLO-World-v2-L & 24.2\% & 31.7\% & 26.3\% & 10.7\% & 24.0\% & 31.8\% & 188.91 M & 308.5 \\
\midrule
Proposed-CSP & 15.3\% & 22.2\% & 16.5\% & 7.0\% & 15.4\% & 19.6\% & 63.20 M & 158.5 \\
Proposed-MSP & 22.4\% & 30.8\% & 24.3\% & 11.6\% & 22.9\% & 28.5\% & 102.61 M & 373.0 \\
\bottomrule
\end{tabular}
\end{table*}

\subsection{Text Processing}
This study employs BERT\cite{devlin2018bert} model to handle descriptive text. BERT Transformer architecture processes the text by considering the context of each word, capturing its semantic meaning more precisely. The processed text is encoded into high-dimensional vectors that represent the semantic information and relationships between words. Additionally, WordNetLemmatizer is used to standardize word forms, and relevant keywords are extracted from the text. This structured representation of semantic data enhances the effectiveness of subsequent processing tasks.Use the BERT model to encode sentences, generating corresponding word embedding vectors \textit{\textbf{P}}.

\begin{equation}
P=BERT(T_i)
\end{equation}

After obtaining the list of data processed by the BERT model, the next step involves further processing, including lemmatization using WordNetLemmatizer and keyword extraction. Lemmatization with WordNetLemmatizer aims to reduce words to their base forms or roots, such as converting different verb tenses (e.g., "walking," "walked") to their root form "walk." This process unifies word forms, reduces data redundancy, and simplifies subsequent analysis. Each word in the BERT-processed list, along with its labels and attributes, is lemmatized, replacing the original words with their base forms while preserving their part-of-speech tags.Use WordNetLemmatizer to convert all word embedding vectors into their base form vectors 
\(P_O\).

\begin{equation}
P_O=WordNetLemmatizer(P)
\end{equation}

When lemmatization is performed, the basic form or stem of the original vocabulary will be obtained. Such processing helps normalize different forms of words, making it easier to compare similarities or matches between them. Subsequently, the FastText model was used to calculate the category similarity between the restored words and the category vocabulary database \textbf{C}. The FastText model is a word vector model that calculates the semantic relationship between words by mapping words into a high-dimensional vector space \(C_j\).
\begin{equation}
P_{O^i} \xrightarrow{\text{map}} C_{j^*}, \quad \text{where} \quad
j^* = \arg\max_j \frac{P_{o^i} \cdot C_j}{\|P_{o^i}\| \|C_j\|}
\end{equation}

Finally, the existing vocabulary and the matched vocabulary, after calculating similarity, undergo further vocabulary search. Objects related to these terms are extracted from the text list and matched with the category vocabulary in the database. This process allows for the identification and extraction of specific objects mentioned in the text that are related to the categories in the database. The Encoded objects PI will be used in the subsequent image processing stage.
\begin{equation}
P_I= Compare(P_O,D)
\end{equation}

\subsection{Multimodal Interaction}

Finally, using the previously obtained Image-aware Embedding, filter all categories and bounding boxes output by the image model. This filtering process takes into account feature similarity and semantic information, discarding detection results that do not align with the semantic information. The filtered categories and bounding boxes are then output as the final detection results, including the class labels and precise bounding box coordinates for each object in the image.Semantic Filtering of Image Objects: Based on the object vectors \( P_I \) generated by the Image-aware Embedding, filter all object categories \( C_i \) and bounding boxes \( B_i \) produced by the image model. This results in the final objects \( C_{prompt} \) and \( B_{prompt} \) that are included in the input sentence.

\begin{equation}
C_prompt  ,B_prompt=Filter(C_i,B_i)
\end{equation}

\begin{figure*}[!ht]
    \centering
    \includegraphics[width=0.8\linewidth]{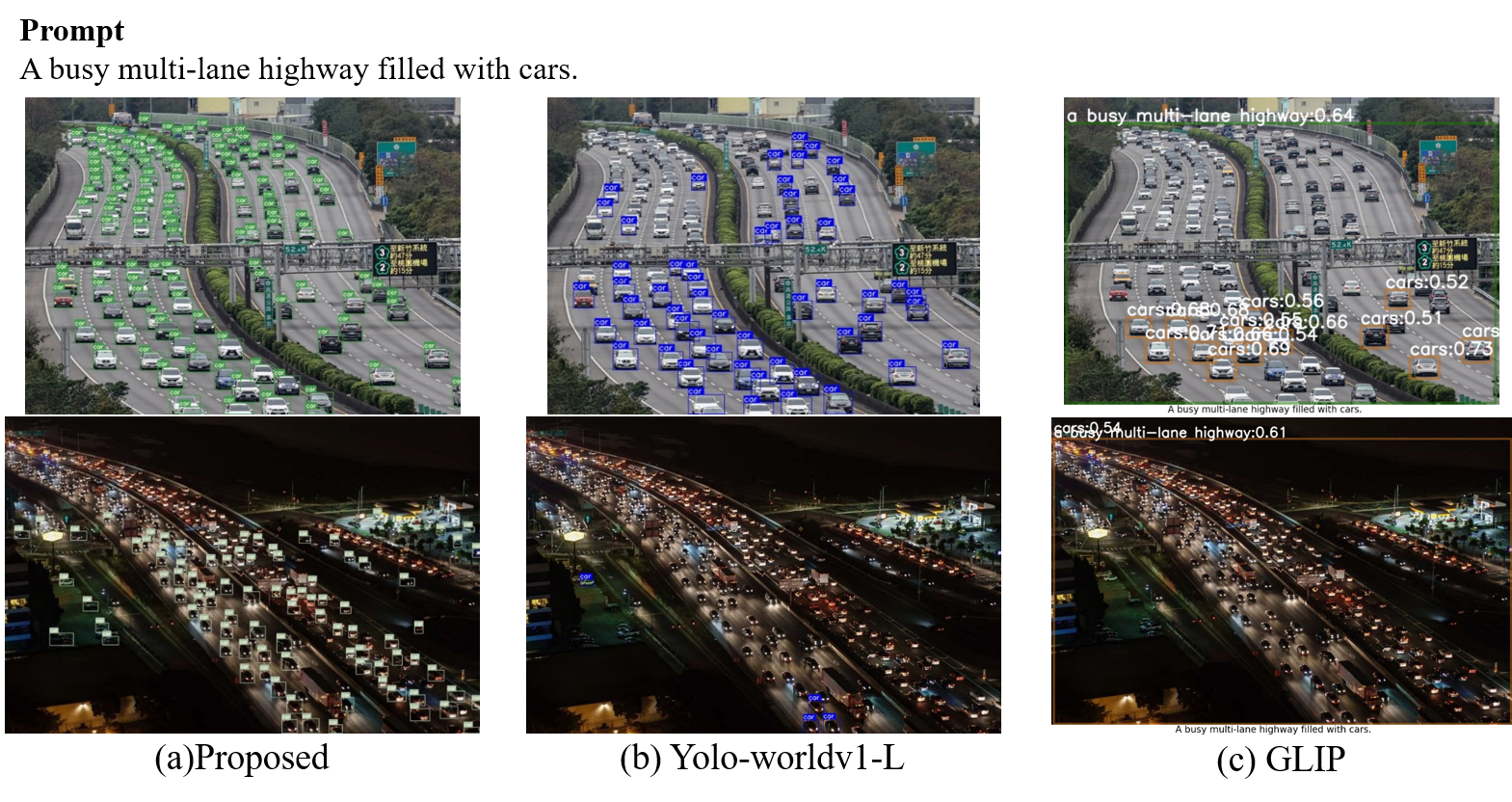}
    \caption{Experimental Results Comparison 1}
    \label{fig:car}
\end{figure*}

\begin{figure*}[!ht]
    \centering
    \includegraphics[width=0.8\linewidth]{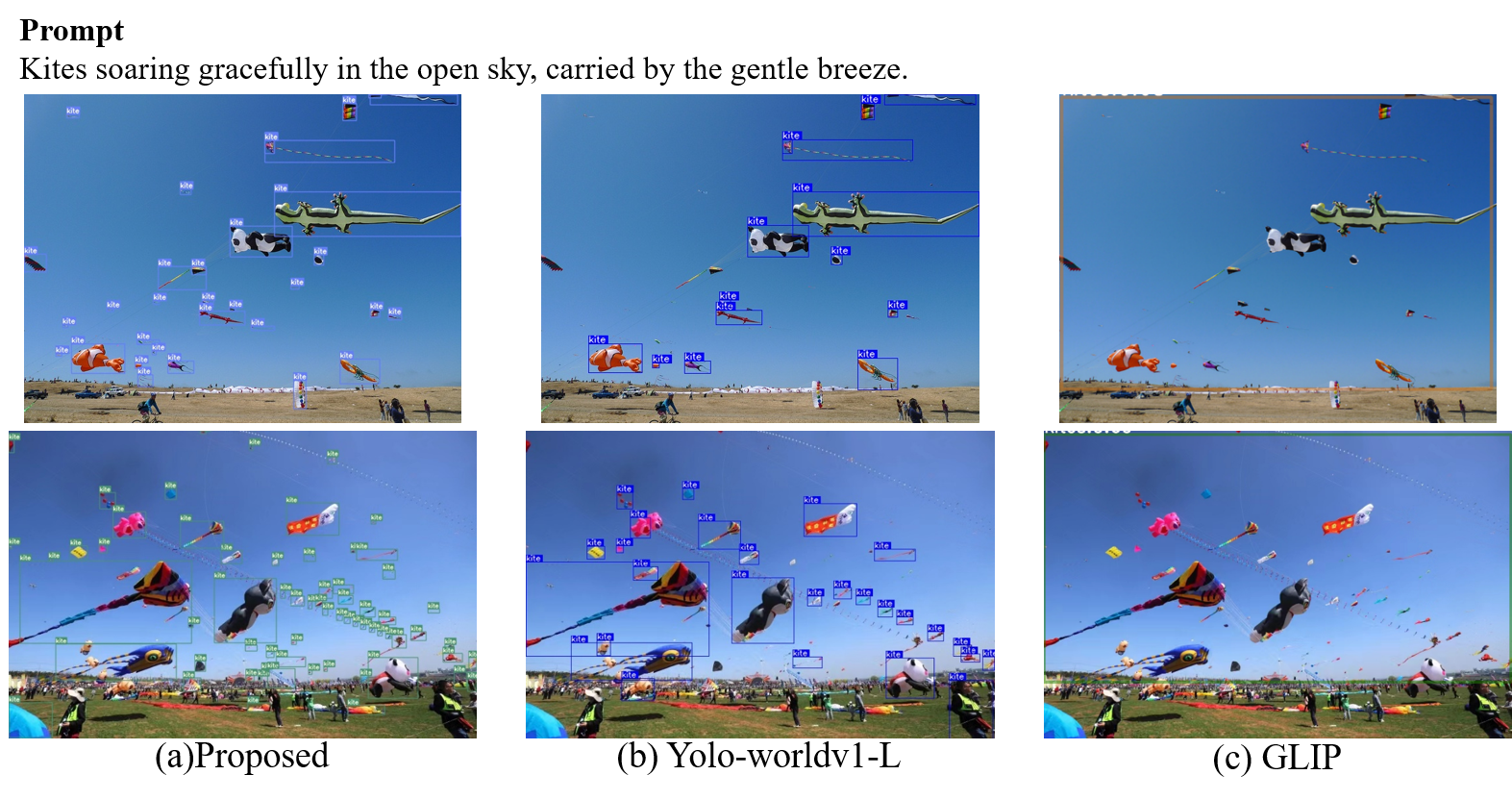}
    \caption{Experimental Results Comparison 2}
    \label{fig:kite}
\end{figure*}

\begin{figure*}[!ht]
    \centering
    \includegraphics[width=0.8\linewidth]{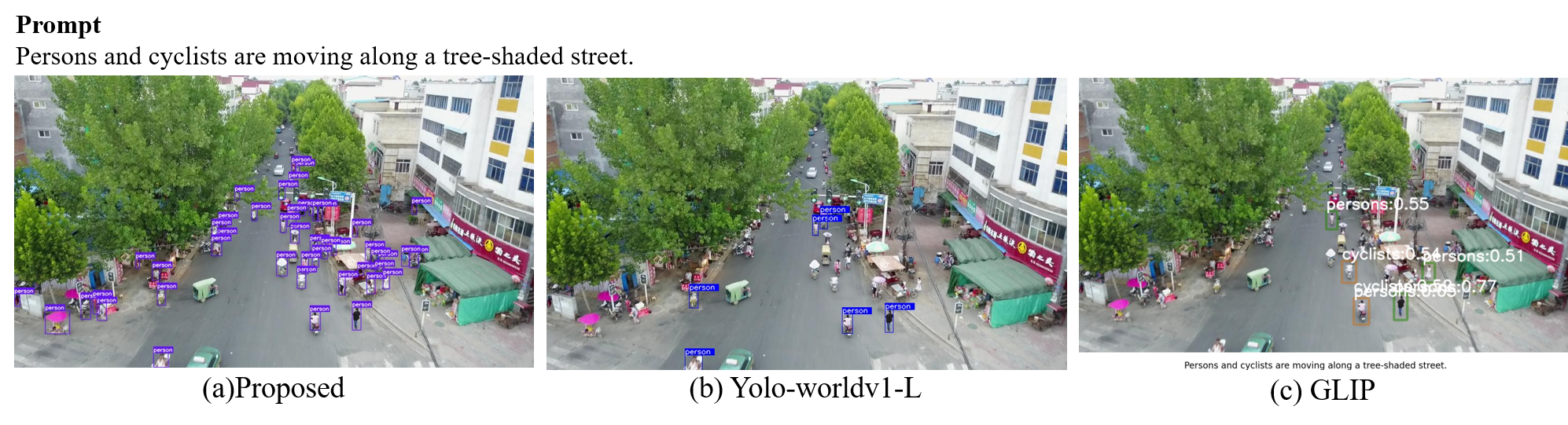}
    \caption{Experimental Results Comparison 3}
    \label{fig:top}
\end{figure*}

\section{Experiments}
\subsection{Dataset}
Two distinct datasets were employed: COCO and Objects365. These datasets differ in terms of the categories and quantity of objects they encompass. Therefore, both COCO and Objects365 datasets were separately used to train the model to leverage their unique features and advantages. This approach allows for better handling of various scenarios and application requirements, enhancing the generalizability and applicability of the research outcomes.

\subsection{Comparison with the State-of-the-art Methods}
\textbf{Performance on COCO.}The advantages are clearly demonstrated in comparisons with other technologies. For instance, compared to YOLO-World\cite{cheng2401yolo}, which also employs a Transformer+CNN architecture, our method significantly outperforms on the COCO2017 validation dataset with a performance of 52.6\%, whereas YOLO-World achieves only 45.8\%. Although our method slightly lags behind GLIP\cite{li2022grounded} in terms of performance, it requires half the number of parameters.

\textbf{Performance on Objects365.}The proposed model achieves an AP of 22.4\%, which is comparable to the YOLO-World-v2-L model's AP of 24.2\%, but with significantly fewer parameters (102.61 M vs. 188.91 M) and lower GFLOPs (373 vs. 308.5).For AP\(_{50}\) and AP\(_{75}\), the proposed model scores 30.8\% and 24.3\% respectively, demonstrating competitive performance against the YOLO-World-v2-L model, which scores 31.7\% and 26.3\%.The proposed model excels particularly in detecting small and medium objects, with AP\(_s\) of 11.6\% and AP\(_m\) of 22.9\%, outperforming most of the other models significantly in these categories. This is crucial for applications requiring precise detection of smaller objects.For large objects, the proposed model's AP\(_l\) of 28.5\% is close to the highest performance of 31.8\% achieved by YOLO-World-v2-L, showing its robustness across different object sizes.The lower parameter count (102.61 M) and moderate GFLOPs (373) indicate that the proposed model is more efficient, making it suitable for deployment in resource-constrained environments without compromising much on detection performance.

\subsection{Visualization of Results Comparison}
To better understand the practical performance of the proposed method, selected images from the test set were analyzed in detail shows in Figure \ref{fig:car}, \ref{fig:kite}, \ref{fig:top}. This detailed analysis highlights the specific application effects of the method. Additionally, comparisons with other studies show that the proposed approach significantly outperforms others in terms of inference results.
\section{Conclusion}
A novel approach for integrating two different modalities for vocabulary-based object detection is proposed without feature fusion. The effectiveness of this method has been validated through design and experiments. The method enhances text input processing and encoding, achieving more accurate tokenization of common object names. Optimizations in the image recognition model successfully reduce memory usage during training while maintaining accuracy. The model demonstrates high efficiency even with limited resources. Compared to YOLO-World, which also uses a Transformer + CNN architecture, the proposed method shows superior performance on the COCO dataset, highlighting its advantages without relying on large-scale data or extensive fine-tuning. Future work could focus on better leveraging potential correlations between different modalities while maintaining model accuracy and processing speed to further improve performance. Additionally, as this study does not delve into deeper semantic recognition, future research could explore integrating additional features such as object location and color attributes to enhance the accuracy and granularity of object detection.

\ifCLASSOPTIONcaptionsoff
  \newpage
\fi

\bibliography{main}
\bibliographystyle{ieeetr}

\end{document}